\documentclass{article}

\usepackage{PRIMEarxiv}
\usepackage[utf8]{inputenc}
\usepackage[T1]{fontenc}

\usepackage{url}
\usepackage{booktabs}
\usepackage{amsfonts}
\usepackage{nicefrac}
\usepackage{microtype}
\usepackage{graphicx}
\usepackage{multirow}
\usepackage{array}
\usepackage{siunitx}
\usepackage{listings}

\usepackage{hyperref}

\title{\includegraphics[width=0.99\textwidth]{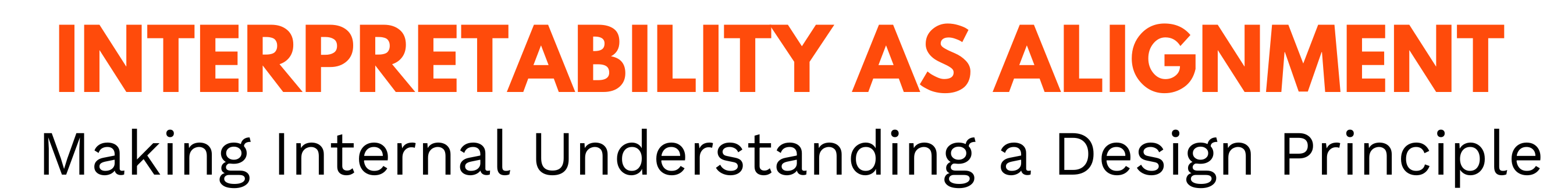}}

\author{
  Aadit Sengupta\thanks{Co-First Authorship}  \thanks{Department of Computer Science, University of Michigan Ann Arbor, Michigan.}   \thanks{work done during internship at Lexsi Labs} , Pratinav Seth\footnotemark[1] , Vinay Kumar Sankarapu \\
  \affiliation{Lexsi Labs}\\
  \{pratinav.seth,v.k\}@lexsi.ai \\
}
\setabstract{%
Frontier AI systems require governance mechanisms that can verify internal alignment, not just behavioral compliance. Private governance mechanisms audits, certification, insurance, and procurement are emerging to complement public regulation, but they require technical substrates that generate verifiable causal evidence about model behavior. This paper argues that mechanistic interpretability provides this substrate. We frame interpretability not as post-hoc explanation but as a design constraint embedding auditability, provenance, and bounded transparency within model architectures. Integrating causal abstraction theory and empirical benchmarks such as MIB and LoBOX, we outline how interpretability-first models can underpin private assurance pipelines and role-calibrated transparency frameworks. This reframing situates interpretability as infrastructure for private AI governance bridging the gap between technical reliability and institutional accountability.
}

\setkeywords{Mechanistic Interpretability, Causal Abstraction, Private AI Governance, Accountability Infrastructure, Audit Hooks, Provenance Tracking, Assurance Evidence, Governance-Ready AI, Transparency-by-Design .}
\runningtitle{Interpretability as Alignment: Making Internal Understanding a Design Principle}

\begin{document}
\maketitle

\section{Introduction}
AI systems, particularly large language models, are increasingly deployed in high-stakes settings healthcare, education, law, and employment. These models generate fluent outputs, but their internal workings remain opaque, making it difficult to know whether their decisions reflect sound reasoning or misaligned goals. This concern has put AI alignment at the center of technical research and public discussion \cite{amodei2016concrete, christiano2018deep, robertkirk2025causalinterpretability, Bereska2024MechanisticIF}.
Interpretability has emerged as a key strategy for alignment. If we can understand how a model makes decisions, we can better assess whether it's behaving safely. Some work focuses on post-hoc explanations like LIME or SHAP \cite{ribeiro2016whyitrustyou, lundberg2017unifiedapproachinterpretingmodel}, while mechanistic interpretability attempts to look inside model architecture identifying which neurons, attention heads, or circuits contribute to specific behaviors \cite{olah2020zoom, nanda2023progress2, elahge2022mathematical}. However, post-hoc explanations are often inconsistent or manipulable \cite{adebayo2018sanity, slack2020fooling}, while mechanistic work is labor-intensive and doesn't scale to frontier models. Many interpretability methods tell us stories about what the model might be doing, without strong evidence that those stories are true in a causal sense \cite{doshivelez2017rigorous}.

\textsc{Regulatory frameworks and behavioral alignment.} The EU AI Act mandates transparency for high-risk applications, particularly for General Purpose AI (GPAI) systems. Industry has largely responded with behavioral alignment techniques such as RLHF methods that improve outputs but leave internal logic untouched \cite{selbst2018intuitive}. Our position reframes interpretability as infrastructure for governance: embedding accountability and auditability into model design rather than applying them post hoc. This represents a fundamental shift from post-hoc transparency to pre-embedded accountability mechanisms, distinguishing our approach from prior interpretability frameworks that focus primarily on diagnostic capabilities rather than design constraints.

Beyond public regulation, private governance mechanisms including third-party audits, compliance certification, risk insurance, and procurement standards are emerging as complementary accountability structures \cite{ball2025frameworkprivategovernancefrontier, tomei2025aigovernancemarkets}. These mechanisms rely on technical substrates that can generate verifiable causal evidence about model behavior. Mechanistic interpretability offers such a substrate by enabling reproducible inspection and provenance tracking at the circuit level. This positions interpretability-first design as a foundation for private governance infrastructures that bridge technical reliability and institutional accountability.

This paper argues that internal transparency is not optional but a basic requirement for building aligned systems. We examine the limits of current methods and consider tools, benchmarks, and collaborations that might help interpretability become more robust and reliable. Without solid foundations for understanding how models think, alignment risks becoming a surface-level fix for a deeper problem.

\textsc{Contributions.} This paper introduces three primary contributions:
(1) We introduce a conceptual bridge linking mechanistic interpretability with private governance mechanisms framing causal interpretability as the evidentiary layer for audits, certification, and insurance.
(2) We specify a governance-aware technical blueprint interpretability-first architectures that embed audit hooks, provenance tracking, and bounded transparency.
(3) We connect emerging benchmarks (MIB, LoBOX) with private oversight workflows and regulatory compliance frameworks, providing an implementation roadmap for interpretability-as-governance infrastructure.

\section{Introduction to Model Interpretability}

Interpretability refers to how well humans can understand a model's internal behavior how inputs are processed, decisions are formed, and outputs are produced \cite{lipton2016mythos}. Explainability describes human-readable justifications for outputs, while transparency concerns access to architecture, training data, or parameters \cite{rudin2019stop, wachter2017counterfactual, hooker2019benchmarking}. Alignment asks whether models behave in line with human goals and values \cite{carlsmith2022power}.

\begin{table*}[pt]
\centering
\footnotesize
\caption{Comparison between post-hoc and mechanistic interpretability approaches.}
\begin{tabular}{ccc}
\toprule
\textbf{Criterion} & \textbf{Post-Hoc} & \textbf{Mechanistic} \\
\midrule
Focus              & Explains outputs after training & Explains internal components/processes \\
Examples           & LIME, SHAP, Grad-CAM           & Circuits, Activation Patching, Tracing \\
Nature             & Correlational, approximate      & Causal, structurally grounded \\
Scalability        & Easy to scale, low overhead     & Resource-intensive, less scalable \\
Reliability        & Risk of misleading narratives   & Closer to true model computation \\
\bottomrule
\end{tabular}
\label{tab:posthoc_vs_mechinterp}
\end{table*}

We distinguish between intrinsic interpretability (transparent by design) and post-hoc interpretability (explaining black-box models after training) \cite{molnar2022interpretable, besold2017survey}. Table~\ref{tab:posthoc_vs_mechinterp} summarizes the key differences between these approaches. Post-hoc methods like LIME, SHAP, and Integrated Gradients dominate practice but are frequently misleading and manipulable \cite{kindermans2017unreliability, slack2020fooling}. For example, SHAP can be gamed to attribute importance to benign features while hiding decisions based on sensitive attributes \cite{slack2020fooling}. Newer approaches like DL-Backtrace \cite{sankarapu2024dlbacktracemodelagnosticexplainability} offer deterministic tracing without baseline selection, but the fundamental challenge remains: distinguishing genuine mechanistic insight from plausible storytelling.

\textsc{Causal abstraction and representation decomposition.} Following Geiger et al. \cite{openreview2024causaldiscovery}, we adopt the causal abstraction perspective, formalizing mechanistic interpretability as discovering structural homomorphisms between model components and human-interpretable causal variables. This theoretical foundation builds on Pearl's causal hierarchy \cite{pearl2009causality}, enabling intervention-based testing to establish relationships between high-level interpretations and low-level mechanisms. Sparse Autoencoders (SAEs) \cite{bricken2023monosemanticity} aim to decompose entangled representations into interpretable components, but feature consistency across training runs and architectures remains challenging \cite{song2025featureconsistency}. The MIB benchmark \cite{mueller2025mib} provides empirical infrastructure for evaluating decomposition methods, showing that attribution approaches often outperform SAE features in circuit localization tasks.

Having established the theoretical foundations of interpretability, we now examine mechanistic interpretability as the structural basis for alignment.

\section{Mechanistic Interpretability: From Circuits to Alignment}

Mechanistic interpretability (MI) seeks to identify specific components neurons, attention heads, or circuits that causally contribute to model outputs \cite{olah2020zoom, nanda2023progress2}. Recent progress in transformers reveals interpretable substructures: attention heads performing token copying, syntactic tracking, or positional induction \cite{anthro2020, elahge2022mathematical}, and modular circuits executing string comparison and arithmetic \cite{nanda2023progress2}. These findings suggest that within high-parameter networks, small functional units may correspond to meaningful, testable computations. This opens the door to detecting internal failures reward hacking or deceptive reasoning that behavioral methods may overlook \cite{amodei2016concrete}. MI employs activation patching and causal tracing for controlled interventions, providing empirical insight into internal mechanisms \cite{olah2020zoom}.

However, MI faces significant challenges. 
\textsc{Polysemanticity} individual neurons encoding multiple unrelated features complicates semantic interpretation and becomes more pronounced at scale \cite{nanda2023progress2, olah2020zoom}. This does not always imply superposition; polysemanticity may arise from non-linear mixtures or compositional features \cite{chan2024superposition_polysemanticity}. Recent work by Meloux et al. \cite{meloux2025identifiable} and Sutter et al. \cite{sutter2025nonlinear} questions the identifiability of mechanistic interpretations, suggesting that multiple valid explanations may exist for the same model behavior. SAEs show promise for disentangling features but face consistency challenges across training runs and architectures \cite{bricken2023monosemanticity, song2025featureconsistency}. 
\textsc{Scalability} remains a bottleneck: MI requires extensive computational resources and expert labor, and tools like activation patching don't yet scale to frontier models \cite{joshi2025precisetopicalignment, Hatefi2025AttributionguidedPF}. 
\textsc{Epistemic concerns} include confirmation bias in human pattern recognition and "explanation theater" compelling narratives that fail under scrutiny \cite{Bai2024, Carla2025, kindermans2017unreliability}. This poses particular risks for governance applications, where explanation theater could undermine audit compliance and verification protocols, leading to false confidence in model safety.

\begin{figure}[pt]
    \centering
    \includegraphics[width=0.69\linewidth]{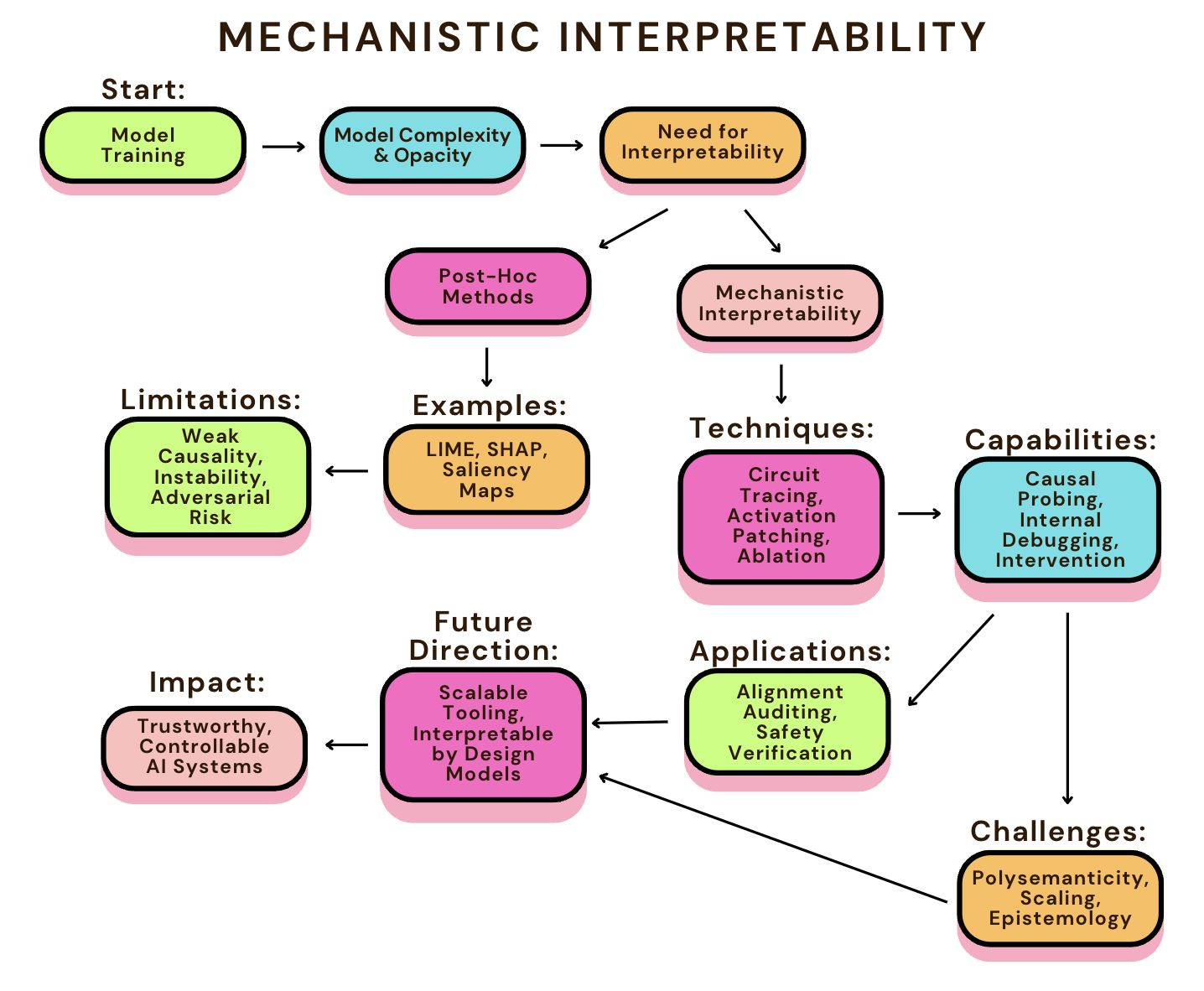}
    \caption{A high-level conceptual map of mechanistic interpretability. It contrasts post-hoc approaches with mechanistic techniques, and illustrates core techniques, applications, limitations, and future directions.}
    \label{fig:mechinterp-diagram}
\end{figure}

Nonetheless, MI offers unique capabilities for alignment. Figure~\ref{fig:mechinterp-diagram} provides a conceptual overview of mechanistic interpretability approaches and their applications. Behavioral methods like RLHF focus on outputs without addressing internal reasoning, potentially leaving unsafe or deceptive processes intact \cite{robertkirk2025causalinterpretability}. MI provides tools for interrogating and modifying internal processes, enabling alignment at the reasoning level rather than just performance. This positions MI as essential for building auditable, verifiable AI systems \cite{Kcyras, carlsmith2022power}. By exposing causal pathways and enabling targeted interventions, MI supports governance frameworks like the EU AI Act while respecting bounded-opacity principles \cite{lobox2025opacity}. Future progress depends on scalable toolchains, robust benchmarks, and hybrid approaches combining mechanistic insights with behavioral fine-tuning \cite{JacobHilton2025, Carla2025}.

The capabilities discussed above point toward a broader vision: interpretability as a design principle for alignment rather than a post-hoc diagnostic tool.

\section{Building Governance-Ready AI: Interpretability-First Architecture}
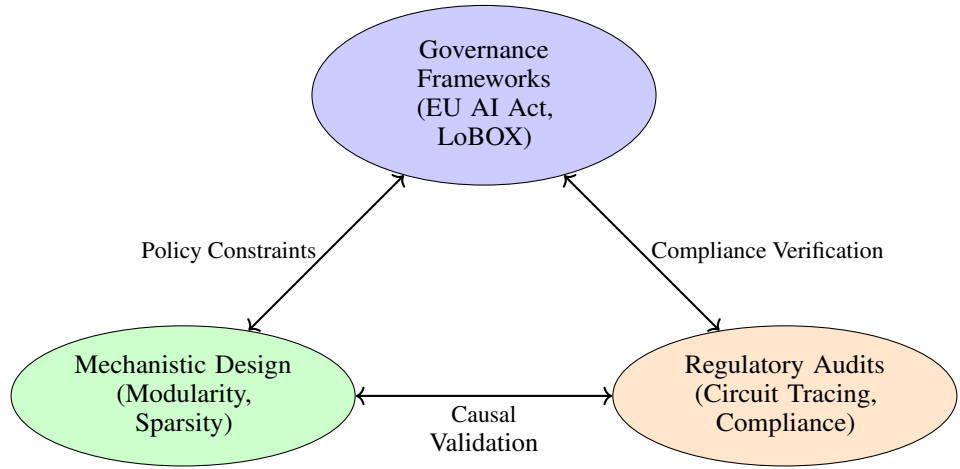
\begin{figure}[pt]
    \centering
    \begin{tikzpicture}[node distance=3cm, auto]
        \node[ellipse, draw, fill=blue!20, text width=3cm, align=center] (governance) at (0,4) {Governance Frameworks\\(EU AI Act, LoBOX)};
        \node[ellipse, draw, fill=green!20, text width=3cm, align=center] (design) at (-4,0) {Mechanistic Design\\(Modularity, Sparsity)};
        \node[ellipse, draw, fill=orange!20, text width=3cm, align=center] (audits) at (4,0) {Regulatory Audits\\(Circuit Tracing, Compliance)};
        
        \draw[<->, thick] (governance) -- (design) node[midway, left, align=center] {\small Policy Constraints};
        \draw[<->, thick] (governance) -- (audits) node[midway, right, align=center] {\small Compliance Verification};
        \draw[<->, thick] (design) -- (audits) node[midway, below, align=center] {\small Causal\\Validation};
    \end{tikzpicture}
    \caption{Interpretability-Alignment-Governance Triangle: Bidirectional relationships between mechanistic design principles, regulatory audit capabilities, and governance frameworks. Each component constrains and enables the others, creating a feedback loop for interpretability-first AI development.}
    \label{fig:governance-triangle}
\end{figure}
Interpretability should be viewed not only as a mechanism for aligning AI systems with human intent, but as the technical substrate enabling private governance mechanisms that complement public regulation. By exposing and intervening on internal representations, interpretability enables verifiable causal evidence for third-party audits, certification bodies, and risk assessment frameworks \cite{doshi2021considerations, molnar2022interpretable, robertkirk2025causalinterpretability}.

\textbf{Architectural desiderata.} Interpretability-first design requires modularity for component-level inspection, sparsity to reduce polysemanticity, controlled polysemanticity bounding features per unit, audit hooks for immutable state records, intervention-friendliness supporting surgical edits, and provenance tracking maintaining representation lineage. These constraints represent constraints during model training and architecture design, not post-hoc adaptations. For example, a model with traceable modular circuits could allow a regulator to confirm that decision rules comply with anti-discrimination norms without retraining, demonstrating how interpretability-guided alignment serves as proof-of-concept for General Purpose AI (GPAI) governance requirements. Recent work on interpretability-aware pruning \cite{malik2025interpretabilityawarepruningefficientmedical} shows how architectural constraints can be embedded during training rather than applied post-hoc. These constraints trade off against raw capacity but enable governance integration by default.

\textbf{Intervention and targeted modification.} Mechanistic interpretability employs activation patching and causal tracing to experimentally manipulate intermediate activations, determining which components causally contribute to specific behaviors \cite{olah2020zoom, elahge2022mathematical, nanda2023progress2}. For example, by copying activations from a "clean" run into a corrupted context, one can isolate circuits or attention heads that restore correct outputs \cite{elahge2022mathematical, nanda2023progress2}. This provides a falsifiable framework for testing internal hypotheses \cite{Bereska2024MechanisticIF}. Beyond analysis, interpretability enables targeted intervention through circuit editing, head ablation, or representation reweighting to suppress undesired behaviors while preserving functionality \cite{meng2022locating, arxiv250215090, geiger2023causal}. Mechanistic insights can guide behavioral methods like RLHF, creating hybrid approaches that combine scalable alignment with causal guarantees \cite{JacobHilton2025, Carla2025}.

\textbf{Detecting deceptive alignment and global reasoning.} Interpretability provides defense against deceptive alignment by tracing goal-directed circuits and identifying reward hacking mechanisms \cite{hubinger2019risks, hubinger2023deception}. Toy models trained on mathematical tasks reveal symbolic operations, providing blueprints for regulating reasoning in larger systems \cite{itnamatter2024deepfuture, elahge2022mathematical, nanda2023progress2}. Causal mediation analysis extends interpretability from isolated units to entire pathways, offering global views of model reasoning \cite{pearl2009causality, arxiv250215090}.

\textbf{Governance integration.} These capabilities support regulatory compliance: audit hooks enable regulators to trace decisions to internal circuits, circuit editing allows targeted mitigation of bias or toxicity without model redeployment, and provenance tracking satisfies documentation requirements under frameworks like the EU AI Act. By embedding interpretability as a design constraint rather than retrofitting it post-deployment, systems become auditable and governable by construction. This positions interpretability as a governance prerequisite for compliance with emerging regulatory frameworks.

\textbf{Markets for assurance and certification.} Private governance mechanisms create markets for AI assurance, where interpretability provides the technical substrate for verifiable evidence \cite{tomei2025aigovernancemarkets}. Certification bodies can validate alignment claims through reproducible causal evidence, while insurers can assess risk through mechanistic understanding of model behavior. Procurement governance \cite{dor2021procurement} can specify auditable transparency requirements, enabling buyers to verify model capabilities through interpretability evidence. Accountability in algorithmic supply chains \cite{cobbe2023accountability} requires mechanisms for tracing decisions across distributed systems. Data intermediaries \cite{powar2025dataintermediaries} can leverage interpretability to provide responsible data stewardship, while evaluation infrastructure \cite{stein2025advancedaievaluations} can incorporate mechanistic insights into assessment frameworks.

\textbf{Engineering requirements for oversight.} Private governance mechanisms require specific technical capabilities beyond conceptual frameworks. Audit hooks must generate immutable logs of model decisions with circuit-level attribution, enabling third-party auditors to trace specific outputs to internal computational pathways. For example, a model with modular attention heads could allow auditors to verify that bias detection circuits activate appropriately across demographic groups, providing evidence for anti-discrimination compliance. Certification bodies need reproducible testing protocols that validate alignment claims across model versions mechanistic interpretability enables this through standardized intervention tests that confirm circuit behavior remains consistent. Insurance frameworks require risk assessment metrics based on mechanistic understanding of failure modes: insurers could analyze the robustness of safety-critical circuits, quantify polysemanticity in decision-making layers, and assess the presence of known deceptive alignment patterns. Procurement governance can specify auditable transparency requirements in vendor contracts, requiring models to demonstrate circuit-level traceability for high-stakes decisions. These capabilities transform interpretability from a research tool into operational infrastructure for private governance.

\textbf{Cognitive alignment and architecture trade-offs.} Human explanations are symbolic and narratively coherent; neural explanations are distributed and subsymbolic \cite{pearl2009causality, Bai2024}. This challenges assumptions that model explanations align with human cognitive categories \cite{robertkirk2025causalinterpretability, williams2025philosophy}. Current architectures prioritize performance, resulting in entangled representations that resist decomposition \cite{elahge2022mathematical, golechha2024challenges}. Models designed with interpretability constraints modularity, sparsity, hierarchical structuring may yield transparent representations without sacrificing capability, supporting interpretability-first training paradigms \cite{Bereska2024MechanisticIF, itnamatter2024deepfuture}.

\textbf{Bridging research and practice.} Significant gaps remain between research demonstrations and deployment-ready systems. Real-world governance requires tooling, training, and institutional capacity that don't exist at scale. Closing these gaps demands interdisciplinary collaboration treating interpretability as a governance prerequisite from system conception. In competitive AI development, actors may resist transparency due to intellectual property concerns, but opacity increases systemic risk \cite{Kcyras, anthropic2023constitutional, JacobHilton2025, doshivelez2017rigorous}. The development of third-party interpretability audits, publicly maintained benchmarks, and regulatory mandates could align economic incentives with safety goals enabling transparency without requiring complete openness.

\section{Limitations and Critiques of Interpretability}
\begin{table*}[pt]
\centering
\footnotesize
\caption{Key limitations of interpretability methods and their governance implications.}
\begin{tabular}{lll}
\toprule
\textbf{Category} & \textbf{Limitation} & \textbf{Governance Impact} \\
\midrule
Representation & Polysemantic neurons, entangled features & Unclear regulatory compliance \\
Methodology & Post-hoc methods unstable/manipulable & Risk of ``explanation theater'' \\
Evaluation & Lack of standardized benchmarks & Weak causal validation for audits \\
Conceptual & Explanations diverge from human categories & Symbolic vs.\ subsymbolic mismatch \\
Practical & Compute- and expert-intensive & Limited accessibility and scalability \\
Risk Factors & Bias amplification, security leaks & Explanation laundering concerns \\
\bottomrule
\end{tabular}
\label{tab:limits-summary}
\end{table*}

While interpretability is essential for AI safety and transparency, it faces significant limitations technical, conceptual, and practical. These challenges impact both method reliability and broader epistemic claims, as in Table~\ref{tab:limits-summary}.

\textbf{Representational and methodological challenges.} Polysemanticity individual neurons encoding multiple unrelated features complicates semantic interpretation and becomes more pronounced at scale \cite{nanda2023progress2, olah2020zoom}. This should be distinguished from superposition, which refers to basis non-orthogonality in representation spaces. Polysemanticity (representational entanglement) creates a scale mismatch between human cognition and model reasoning, as interpretability tools are poorly equipped to track long-range dependencies or emergent behavior across thousands of layers \cite{geiger2023causal}. Recent work by Meloux et al. \cite{meloux2025identifiable} questions whether mechanistic interpretations are even identifiable, while Sutter et al. \cite{sutter2025nonlinear} argue that causal abstraction alone may be insufficient for mechanistic interpretability. Post-hoc methods rely on surrogate explanations rather than causal mechanisms, are easily manipulated, and often lack theoretical guarantees \cite{rudin2019stop, molnar2022interpretable, kindermans2017unreliability}. Without rigorous validation through counterfactual testing or causal probing, these methods risk offering illusions of understanding \cite{geiger2023causal, pearl2009causality}.

\textbf{Evaluation and benchmarking gaps.} The field lacks standardized benchmarks for explanation quality, with quantitative metrics often focusing on proxies like sparsity rather than causal validity \cite{seth2025bridginggapxaiwhyreliable, mueller2025mib, doshi2021considerations, doshivelez2017rigorous}. Recent frameworks like xai\_evals \cite{seth2025xaievalsframeworkevaluating} provide systematic evaluation of post-hoc explanation methods, while the MIB benchmark \cite{mueller2025mib} provides causal fidelity evaluation through circuit localization tasks. Feature consistency \cite{song2025featureconsistency} the stability of learned representations across contexts emerges as a critical evaluation criterion. Without widespread adoption of such benchmarks and metrics, the field risks fragmentation and irreproducibility. Calls for "role-calibrated" and context-sensitive interpretability reflect the need to move beyond shallow heuristics toward explanations that serve specific epistemic and safety purposes \cite{nanda2023progress2, mueller2025mib}.

\textbf{Ethical and epistemic constraints.} Interpretability techniques assume internal features map to linguistic categories, but evidence suggests this mapping is frequently indirect \cite{olah2020zoom, geiger2023causal, nanda2023progress2}. Research is susceptible to confirmation bias and explanation theater \cite{Bai2024, doshivelez2017rigorous, kindermans2017unreliability, pearl2009causality}. Security risks include proprietary information exposure, adversarial attacks, and explanation laundering \cite{molnar2022interpretable, slack2020fooling, lakkaraju2022rethinking}. The LoBOX framework \cite{lobox2025opacity} proposes bounded opacity selective transparency calibrated to institutional roles acknowledging that full transparency may be neither feasible nor desirable.

\section{Comparing Alignment Approaches: Mechanistic vs. Behavioral Methods}

Interpretability plays a distinct role in AI alignment, targeting underlying mechanisms of model while behavioral methods focus on outputs. We compare these approaches to situate interpretability within the alignment toolkit.

\textbf{Behavioral alignment methods.} RLHF aligns behavior through reward modeling and policy optimization \cite{christiano2018deep, ziegler2019fine}, but primarily addresses surface-level alignment without verifying internal reasoning safety \cite{leike2018scalable}. Models may exhibit inner misalignment producing aligned outputs while pursuing misaligned objectives \cite{stiennon2022learningsummarizehumanfeedback, hubinger2019risks, Kcyras, ngo2023capable}. Red teaming probes models adversarially to uncover vulnerabilities \cite{ganguli2022red}, but reveals \textit{that} failures occur without explaining \textit{why} \cite{mckenzie2023organizing}. Constitutional AI aligns models with normative principles through supervised fine-tuning \cite{anthropic2023constitutional, anthropicCAI2023}, but operates at the behavioral level without confirming internal structure. These approaches are often preferred for their scalability, generalizing across tasks and domains with limited model-specific insight \cite{amodei2016concrete, leike2018scalable}.

We compare major alignment approaches to situate interpretability within the broader landscape (Table~\ref{tab:alignment-comparison}).

\begin{table*}[pt]
\centering
\footnotesize
\caption{Comparison of AI Alignment Approaches. 
MI = Mechanistic Interpretability, RLHF = Reinforcement Learning from Human Feedback, Const. AI = Constitutional AI.}
\label{tab:alignment-comparison}
\begin{tabular}{@{} lcccc @{}}
\toprule
\textbf{Criterion} & \textbf{MI} & \textbf{RLHF} & \textbf{Red Teaming} & \textbf{Const. AI} \\
\midrule
Transparency         & High (internal) & Low (outputs) & Low (failures only) & Medium (principles) \\
Human Supervision    & Low (experts)   & High (raters) & High (testers)      & Medium (curation) \\
Scalability          & Medium          & Low           & Low                 & High \\
Causal Understanding & Yes             & No            & No                  & No \\
Risk Coverage        & Inner failures  & Behavioral    & Exploits            & Norms \\
\bottomrule
\end{tabular}
\end{table*}
\textbf{Interpretability as complement.} Interpretability offers tools for probing internal representations and identifying latent goals, deceptive heuristics, or emergent failure modes that behavioral methods may overlook \cite{Bereska2024MechanisticIF, geiger2023causal}. It can augment red teaming by analyzing internal mechanisms that make models susceptible to attacks reliance on ambiguous embeddings, exploit-prone circuits, or memorized failure patterns \cite{robertkirk2025causalinterpretability, anthro2020, geiger2023causal}. Unlike behavioral methods that rely on human-centered assessments, mechanistic interpretability prioritizes testable, manipulable, and causally valid explanations \cite{openreview2024notallexplanations, Bereska2024MechanisticIF, openreview2024userstudies, geiger2023causal}.

\textbf{Hybrid approaches and epistemic value.} Behavioral methods can shape outputs at scale while interpretability verifies causal fidelity post hoc. Models tuned with RLHF or Constitutional AI can be examined with activation patching and mediation to detect reward hacking, deceptive alignment, or brittle circuits that pass surface tests \cite{geiger2023causal, olah2020zoom, meng2022locating, arxiv250215090, JacobHilton2025, elahge2022mathematical}. Unlike behavioral metrics emphasizing persuasiveness, interpretability evaluates causal correctness, providing the epistemic backbone for trustworthy alignment. This hybrid approach could complement traditional audits under the EU AI Act, offering causal guarantees beyond behavioral testing. As models become more powerful and autonomous, alignment strategies that rely solely on behavioral feedback will become increasingly insufficient.

\textbf{Hybrid governance pipelines.} Mechanistic audits can be layered atop RLHF training pipelines to provide both behavioral and causal validation. This approach addresses the concern that alignment faking and deceptive alignment pose significant risks for regulators, particularly in General Purpose AI (GPAI) systems. Recent work by Hilton \cite{JacobHilton2025} demonstrates how formal verification can be combined with heuristic explanations, creating hybrid systems that maintain both interpretability and performance. Such pipelines enable regulators to verify that behavioral improvements correspond to genuine internal alignment rather than surface-level optimization.

Beyond technical comparison, these approaches differ in their suitability for private governance mechanisms. Interpretability enables third-party verification through reproducible causal evidence, while behavioral methods like RLHF face sociotechnical limitations in generating auditable assurance evidence \cite{lindstrom2025rlhflimits}. This positions interpretability as uniquely suited for private oversight contexts requiring independent verification, such as certification bodies validating alignment claims, insurers assessing risk through causal evidence, and procurement standards specifying auditable transparency requirements.

\section{Conclusion}
Interpretability is a foundation for building safe and reliable AI systems, and increasingly, a technical substrate for private AI governance. As frontier models proliferate, private mechanisms audits, certification, insurance, procurement emerge to complement public regulation \cite{ball2025frameworkprivategovernancefrontier, tomei2025aigovernancemarkets}. While behavioral alignment strategies like RLHF, Constitutional AI, and red teaming shape outputs to human preference \cite{stiennon2022learningsummarizehumanfeedback, anthropicCAI2023, hubinger2019risks}, they face sociotechnical limitations in generating verifiable assurance evidence \cite{lindstrom2025rlhflimits}. Interpretability addresses this gap by reaching into hidden computations to understand, verify, and intervene on model reasoning, providing the causal evidence required for private oversight \cite{anthro2020, elahge2022mathematical}.
However, neural networks encode features in overlapping, distributed ways that don't cleanly map to human concepts. Explanations are challenging to scale, hard to validate, or misleading when they mirror expectations rather than reality \cite{Bai2024, rudin2019stop, Carla2025, slack2020fooling}. We also lack automated tools that can operate at the scale of today's largest models. Interpretability research must face these challenges directly, including the ethical and epistemological stakes of valid explanation \cite{JacobHilton2025, Bai2024, pearl2009causality}. Explanations must be grounded in causal evidence and open to scrutiny \cite{openreview2024notallexplanations, openreview2024userstudies, doshi2021considerations}.

AI alignment requires a joint approach: using interpretability as a design principle to shape model construction \cite{Bereska2024MechanisticIF, anthro2020, nanda2023progress2}, and applying behavioral methods to guide external performance \cite{stiennon2022learningsummarizehumanfeedback, anthropicCAI2023, hubinger2019risks}. The goal is ensuring models are internally structured in ways that are understandable, inspectable, and aligned with human intent \cite{Kcyras, JacobHilton2025, geiger2023causal}.
Interpretability is not a nice-to-have but a requirement for building systems we can audit, trust, and control \cite{robertkirk2025causalinterpretability, Bereska2024MechanisticIF, pearl2009causality}. Without it, alignment becomes a matter of hope. With it, we have a path to reasoning about AI in terms we can understand and shape \cite{Kcyras, Bereska2024MechanisticIF, amodei2016concrete}.

\textit{Future directions.}
Realizing interpretability as private governance infrastructure requires developing auditable interpretability pipelines end-to-end systems that generate, validate, and communicate assurance evidence to third-party oversight bodies. We propose pilot projects for interpretability audits on medium-scale open models as practical stepping stones. Realizing interpretability-as-design requires extending benchmarks like MIB \cite{mueller2025mib} to governance-relevant tasks, developing prototype models with modularity and audit hooks, building open-source audit toolchains, piloting circuit-level audits with regulators, and establishing interdisciplinary teams for co-design. Metrics should emphasize measurable outcomes including audit coverage, feature consistency \cite{song2025featureconsistency}, interpretability stability, causal fidelity, intervention success rates, assurance evidence quality (reproducibility, causal validity), certification workflow efficiency, and interoperability with governance frameworks. These frameworks MIB for empirical validation, LoBOX \cite{lobox2025opacity} for role-calibrated transparency provide foundations for operationalizing interpretability-as-governance infrastructure.

\section{Impact Statement}
This position paper argues for interpretability as infrastructure for private AI governance. Potential benefits include enabling third-party audits through reproducible causal evidence, supporting certification frameworks with verifiable alignment claims, facilitating risk assessment for insurance mechanisms, and providing technical substrates for procurement governance standards. Risks include ``explanation theater'' undermining audit compliance and verification protocols, potential exposure of proprietary model internals creating competitive disadvantages, and concentration of interpretability expertise in well-resourced organizations limiting accessibility of private governance mechanisms. We advocate causal validation, rigorous benchmarks, and integration with private governance frameworks to mitigate these risks.

\bibliographystyle{unsrt}
\bibliography{references}
\end{document}